%% file: cvpr_5767_final.tex
\documentclass[10pt,twocolumn,letterpaper]{article}

\usepackage{cvpr}
\usepackage{times}
\usepackage{epsfig}
\usepackage{graphicx}
\usepackage{amsmath}
\usepackage{amssymb}
\usepackage{algorithm}
\usepackage{algorithmic}
\usepackage{multirow}
\usepackage{subfig} 
\usepackage[pagebackref=true,breaklinks=true,letterpaper=true,colorlinks,bookmarks=false]{hyperref}

\cvprfinalcopy % *** Uncomment this line for the final submission\

\newcommand{\bW}{{\mathbf{W}}}
\newcommand{\bc}{{\mathbf{c}}}
\newcommand{\bx}{{\mathbf{x}}}

\newcommand{\bm}{{\mathbf{m}}}
\newcommand{\bb}{{\mathbf{b}}}
\newcommand{\ba}{{\mathbf{a}}}
\newcommand{\bv}{{\mathbf{v}}}
\newcommand{\bA}{{\mathbf{A}}}
\newcommand{\bM}{{\mathbf{M}}}

\def\ie{\emph{i.e.~}}
\def\eg{\emph{e.g.~}}

\newcommand{\zhuo}[1] {{\color{black} {#1}}}  
\newcommand{\zhuonew}[1]{{\color{black} {#1}}}

\def\cvprPaperID{5767} % *** Enter the CVPR Paper ID here

\ifcvprfinal\fi
\begin{document}
	
	%%%%%%%%% TITLE
	\title{Cogradient Descent for Bilinear Optimization}

	\author{Li'an~Zhuo,\textsuperscript{1} Baochang~Zhang,\textsuperscript{1}\thanks{Baochang Zhang is the corresponding author.} \ Linlin~Yang,\textsuperscript{2} Hanlin~Chen,\textsuperscript{1} Qixiang~Ye,\textsuperscript{3} \\David~Doermann,\textsuperscript{4} Guodong~Guo,\textsuperscript{5} Rongrong~Ji, \textsuperscript{6}\\
		\textsuperscript{1} School of Automation Science and Electrical Engineering, Beihang University, \\ 
		\textsuperscript{2} University of Bonn, 
		\textsuperscript{3} University of Chinese Academy of Sciences \\ 
		\textsuperscript{4} University at Buffalo,
		\textsuperscript{5} Institute of Deep Learning, Baidu Research,\\ 
		\textsuperscript{6} School of Information Science and Engineering, Xiamen University\\
		%China\\
		{\tt\small \{lianzhuo, bczhang\}@buaa.edu.cn}\\
		%{\tt\small qxye@ucas.ac.cn\ \ \ doermann@buffalo.edu\ \ \ guoguodong01@baidu.com\ \ \ rrji@xmu.edu.cn} 
}
	
	\maketitle
	\thispagestyle{empty}
	
	%%%%%%%%% ABSTRACT
	\begin{abstract}		
		Conventional learning methods simplify the bilinear model by regarding two intrinsically coupled factors independently, which degrades the optimization procedure. 
		One reason lies in the insufficient training due to the asynchronous gradient descent, which results in vanishing gradients for the coupled variables. 
		In this paper, we introduce a Cogradient Descent algorithm (CoGD) to address the bilinear problem, based on a theoretical framework to coordinate the gradient of hidden variables via a projection function. 
		We solve one variable by considering its coupling relationship with the other,  leading to a synchronous gradient descent to facilitate the optimization procedure.
		Our algorithm is applied to solve problems with one variable under the sparsity constraint, which is widely used in the learning paradigm. We validate our CoGD considering an extensive set of applications including image reconstruction, inpainting, and network pruning. Experiments show that it improves the state-of-the-art by a significant margin\footnote{Source code is available at https://github.com/bczhangbczhang.}. 
	\end{abstract}
	
	%%%%%%%%% BODY TEXT
	\section{Introduction}
	
	Bilinear models are cornerstones of many computer vision algorithms since often the optimized objectives or models are influenced by two or more hidden factors  which interact to produce our observations. 
	With bilinear models we can disentangle for example, illumination and object colors in color constancy, the shape from shading, and object identity and its pose in recognition. 
	Such models have shown great potential in extensive low-level applications including debluring \cite{Young2019solving}, denosing \cite{abdelhamed2019noise}, and $3$D object reconstruction \cite{del2011bilinear}. 
	They have also evolved in convolutional neural networks (CNNs) to model feature interactions, which are particularly useful for fine-grained categorization and  model pruning \cite{liao2019squeezed, liu2017Learning}. 
	
	A basic bilinear problem attempts to optimize the following object function as
	\begin{equation}
	\label{eq:bl_sparse}
	\mathop {\arg \min }\limits_{\bA,\bx} G(\bA,\bx) = \|\bb - \bA\bx\|_2^2\\+\lambda\|\bx\|_1+R(\bA),
	%G(A,\bx^T) = \|b - A\bx^T\|_2^2\\+\lambda_1\|\bx^T\|_1.
	\end{equation}
	where $\bb \in \mathbb{R}^{M \times 1}$ is an observation that can be characterized by $\bA\in \mathbb{R}^{M \times N}$ and $\bx \in \mathbb{R}^{N \times 1}$. \zhuo{$R(\cdot)$ represents the regularization which is normally $\ell_1$ or $\ell_2$ norm.} $\|\bb - \bA\bx\|_2^2$ can be replaced by any function  with the part $\bA\bx$.
	The bilinear models are generally with one variable of a sparsity constraint such as $\ell_1$  regularization, which is widely used in machine learning. The  term contributes to parsimony and avoids overfitting.
	
	Existing methods tend to split the bilinear problem into easy sub-problems, and then solve them using the alternating direction method of multipliers (ADMM). Fourier domain approaches~\cite{bristow2013fast,wohlberg2014efficient} are also exploited to solve the $\ell_1$ regularization sub-problem via soft thresholding. Furthermore, recent works ~\cite{heide2015fast,yang2017image} split the objective into a sum of convex functions and introduces ADMM with proximal operators to speed up the convergence.
	Although the generalized bilinear model~\cite{yokoya2012generalized} considers a nonlinear combination of several end members in one matrix (not two bilinear matrices), it only proves to be effective in unmixing hyperspectral images. These approaches actually simplify the bilinear problems by regarding the two factors as independent, \ie, they optimize  a variable while keeping the other fixed. %{\color{red}  
	
	\begin{figure*}[!t]
		\centering
		\includegraphics[scale=0.80]{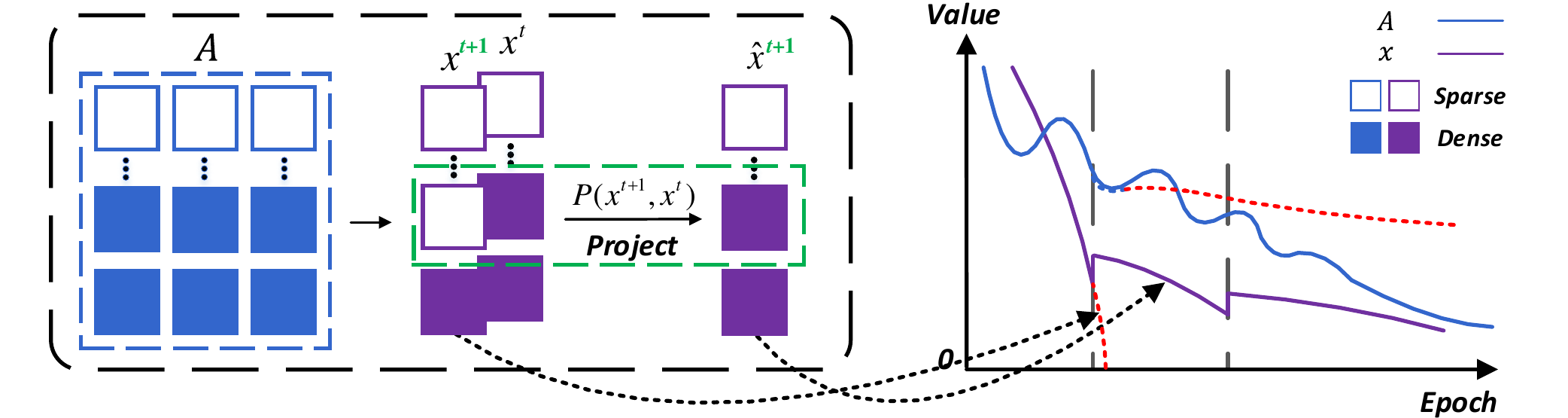}%
		\caption{Illustration of CoGD. Conventional gradient based algorithms has a heuristic that the two hidden variables in bilinear models are independent. Nevertheless, we validate that such an heuristic is implausible and the algorithms suffer from asynchronous convergence and sub-optimal solutions. The red dotted lines denote that the sparsity of $\bx$ causes an inefficient training of $\bA$.  We propose a projection function to coordinate the gradient of the hidden variables.} 
		\label{csc1}
		\vspace{-0.4cm}
	\end{figure*}
	
	Without considering the relationship between two hidden factors, however, existing methods suffer from sub-optimal solutions caused by an asynchronous convergence speed of the hidden variables. 
	As shown in Fig.~\ref{csc1}, when $\bx$ goes sparse with zero values, it causes gradient vanishing for the corresponding element in $\bA$ (see in Eq. \ref{coupled2}). The process results into an insufficient optimization, considering that a corruption in $\bA$ causes { different coding residuals and coding coefficients} for the collaborative representation nature of Eq.~\ref{eq:bl_sparse}~\cite{heide2015fast,zhang2012collaborative}. 
	
	In this paper, we propose a Cogradient Descent algorithm (CoGD) for bilinear models, and target at addressing the gradient varnishing problem by considering the coupled relationship with other variables.
	%We calculate the gradient for one variable by considering its coupled relationship with other variables to address the gradient varnishing problem. 
	%
	A general framework is developed to show that the coupling among hidden variables can be used to coordinate the gradients based on a specified projection function. 
	We apply the proposed algorithm to the problems with one  variable of the sparsity constraint, which is widely used in the learning paradigm. The contributions of this work are summarized as follows
	
	%the nature behind Eq. \ref{eq:bl_sparse} lies in
	
	\begin{itemize}
		\item
		We propose a Cogradient Descent algorithm (CoGD) to optimize gradient-based learning procedures, as well as develop a theoretical framework to coordinate the gradient of hidden variables based on a projection function.  
		
		\item
		%\LY{Our proposed algorithm is highly flexible and efficient for both low-level and deep learning frameworks with bilinear form. To the best of our knowledge, this is the first 
		We propose an optimization framework which decouples the hidden relationship and solves the asynchronous convergence in bilinear models with gradient-based learning procedures.
		
		%The proposed algorithm is easily applied on the most bilinear problems, including convolutional sparse coding and deep learning with bilinear form, based on an unified framework, which decouples the hidden relationship and solves asynchronous convergence efficiently.

		\item
		Extensive experiments demonstrated that the proposed algorithm achieves significant performance improvements on convolutional sparse coding (CSC)  and network pruning. 
	\end{itemize}

	\section{Related Work}
	In this section, we highlight the potential of Eq.~\ref{eq:bl_sparse} and show its applications.  Convolutional sparse coding (CSC) is a classic bilinear problem and has been exploited for image reconstruction and inpainting. Existing algorithms for CSC tend to split the bilinear problem into subproblems, each of which is iterativly solved by ADMM. Furthermore, bilinear models are able to be embedded in CNNs. One application is network pruning. With the aid of bilinear models, we can select important feature maps and prune the channels~\cite{liu2017Learning}. To solve bilinear models in network pruning, some iterative methods like modified Accelerated Proximal Gradient algorithms (APG)~\cite{Huang2017Data} and iterative shrinkage-thresholding algorithms (ISTA)~\cite{ye2018rethinking,Lin2019Towards} are introduced. 
	Other applications like fine-grained categorization~\cite{lin2015bilinear,li2017factorized}, visual question answering (VQA)~\cite{yu2017multi} or person re-identification~\cite{suh2018part}, attempt to embed bilinear models within CNNs to model the pairwise feature interactions and fuse multiple features with attention. To update the parameters in a bilinear model, they directly utilize the gradient descent algorithm and back-propagate the gradients of the loss.

	\section{The Proposed Method}
	
	Unlike previous bilinear models which optimize one variable while keeping another fixed, our strategy  considers the  relationship of  two variables with the benefits on the linear inference. In this section, we first discuss gradient varnishing for bilinear models and then propose our CoGD.
	
	\subsection{Gradient Varnishing}
	
	\noindent Given that $\bA$ and $\bx$ are independent,  the conventional gradient descent method can be used to solve the bilinear problem as
	\begin{equation}
	\bA^{t+1} = \bA^{t} + \eta_1 \frac{\partial G}{\partial \bA},
	\end{equation}
	and
	\begin{equation}
	\label{coupled2}
	%\frac{\partial G}{\partial A} = x^{T,t}\hat{G}(A,x),%2(Axx^T-bx^T).
	(\frac{\partial G}{\partial \bA})^T = \bx^t(\bA\bx^{t}-\bb)^T  = \bx^t\hat{G}(\bA,\bx),%2(Axx^T-bx^T).
	\end{equation}
	where $\hat{G}$ is obtained by considering the bilinear problem  as in Eq.~\ref{eq:bl_sparse}, and we have $\hat{G}(\bA,\bx) = (\bA\bx^{t}-\bb)^T$. It  shows that the gradient for $\bA$ is vanishing when $\bx$ becomes zero, which causes an asynchronous convergence problem. Note that for simplicity, the regularization term is not considered. Similar for $\bx$, we have
	\begin{equation}
	\label{bch1}
	\bx^{t+1} = \bx^{t} + \eta_2 \frac{\partial G}{\partial \bx}.
	\end{equation}
	Here, $\eta_1$, $\eta_2$ represent the learning rate. 
	The conventional gradient descent algorithm for bilinear models iteratively optimizes one variable while keeping another fixed, failing to consider the relationship of the two hidden variables in optimization.
	
	\subsection{Cogradient Descent Algorithm}
	\label{sec:theory}
	We then consider the problem from a new  perspective such that $\bA$ and $\bx$ are coupled. % with the form $\bA=F(\bx)$. 
	It is reasonable that $\bx$ is used to select the element from $\bA$, which are interactive for a better representation. % \zhuo{For convenience, we directly use the norm regularization as $R(A)$.}
	%only when $A$ is sparse
	Firstly, based on the chain rule~\cite{petersen2008matrix} and its notations, we obtain
	
	\begin{equation}
	\begin{aligned}
	\label{coupled1}
	\hat{x}^{t+1}_j &= {x}^{t}_j + \eta_2(\frac{\partial G}{\partial x_j} + Tr((\frac{\partial {G}}{\partial \bA})^T \frac{\partial {\bA}}{\partial x_j})), \\
	\end{aligned}
	\end{equation}
	% 	\begin{equation}
	% 	\frac{\partial {G(\bA)}}{\partial \bx}=\frac{\partial {G(\bA)}}{\partial x_j}=,
	% 	\end{equation}
	where 
	$(\frac{\partial G}{\partial \bA})^T = \bx^t\hat{G}(\bA,\bx)$ as shown in Eq. \ref{coupled2}. $Tr(\cdot)$ represents the trace of matrix, \zhuonew{which means each
	element in matrix $\frac{\partial G}{\partial x_j}$ adds the trace of the corresponding matrix related to $x_j$.}
	Remembering  in Eq. \ref{bch1} that $\bx^{t+1} = \bx^{t} + \eta_2 \frac{\partial G}{\partial \bx}$, Eq.~\ref{coupled1} becomes 
	\begin{equation}
	\label{coupled4}
	%\hat{x}^{T,t+1} = x^{T,t+1} +  \eta_2 x^{T,t} \hat{G}(A,x^T)\frac{\partial A}{\partial {x^T}} ,
	\hat{\bx}^{t+1} = \bx^{t+1} +  \eta_2 \bc^{t} \odot \bx^{t},
	\end{equation}
	where $\odot$ represents the Hadamard product, and $\bc^t = [\sum_i^M\hat{g}_i\frac{\partial \bA_{i1}}{\partial x_1},\dots,\sum_i^M\hat{g}_i\frac{\partial \bA_{iN}}{\partial x_N}]^T$.   $\hat{G}=[\hat{g}_1,\dots,\hat{g}_M]$. 
	 The full derivation of Eq. \ref{coupled4} is given in the supplementary file. It is then  reformulated as a  projection function as
	\begin{equation}
	\label{coupled5}
	\hat{\bx}^{t+1} = P(\bx^{t+1},\bx^{t})=\bx^{t+1} +  \beta \odot  \bx^{t},
	\end{equation}
	where $\beta$ is a parameter   to optimize the variable estimation. It is more flexible than Eq. \ref{coupled1}, because     $\frac{\partial {G}}{\partial \bA}$ becomes varnishing   and $\frac{\Delta \bA}{\Delta x_i}$ is unknown. Our method can be also used to solve the   asynchronous convergence problem,  by controling $\beta$  for the variable estimation. To do that, we first determine when  an asynchronous convergence happens in the optimization based on \zhuonew{a form of logical operation as}
	\begin{equation}
	\label{bch2}
	(\neg s(\bx))\wedge(s(\bA))=1,
	\end{equation}
	and
	\begin{equation}
	\label{sign_function}
	s(*)=
	\begin{cases}
	1& if\ R(*)>\alpha,\\
	0& otherwise,
	\end{cases}
	\end{equation}
	where $\alpha$ represents the threshold which variates for different variables. \zhuo{Eq. \ref{bch2} describes our assumption that an asynchronous convergence  happens for $\bA$ and $\bx$, when their norms become very different. } 
	Accordingly, we  define the update rule of our CoGD as
	\begin{equation}
	\hat{\bx}^{t+1}=
	\begin{cases}
	P(\bx^{t+1},\bx^{t})& if\ (\neg s(\bx))\wedge(s(\bA))=1,\\
	%P(x^{T,t+1},x^{T,t})& otherwise.
	\bx^{t+1}& otherwise,
	\end{cases}
	\label{bt_x}
	\end{equation}
	which leads to an synchronous convergence and  generalizes the conventional gradient descent method.  Then our CoGD is well established.

	% \beta and $\frac{\partial \bA}{\partial {x_j}} =  \frac{\Delta \bA}{\Delta x_j}$, where $\Delta$ denotes the difference of variable over the epoch. $\frac{\partial \bA}{\partial {x_j}} =\mathbf{1}$, if $\Delta x_j$ or $x_j$  approaches to be zero.
	%Accordingly, once the norm of $\bA$ is small enough, $\frac{\partial \bA}{\partial x_j}$ will be approximately equal to $0$, and $\bA$ and $\bx$ can be treated as two independent variables. In this situation,
	\section{Applications}
	We apply the proposed algorithm on CSC and CNNs to validate its performance on image reconstruction, inpainting, and CNNs pruning.
	%We apply our algorithm on CSC and CNNs to validate its performance on image reconstruction and inpainting, and CNNs pruning.
	
	\subsection{CoGD for Convolutional Sparse Coding}
	Unlike patch based sparse coding~\cite{mairal2010online}, CSC operates on the whole image, thereby decomposing a more global dictionary and set of features based on considerably more complicated process. CSC is formulated as
	\begin{equation}
	\begin{aligned}
	\mathop {\arg \min }\limits_{\bA,\bx} 
	&\frac{1}{2}\left\|\bb - \bA\bx \right\|^2_F+\lambda\left\| \bx \right\|_1\\
	\textit{s.t.} &\left\| \ba_k \right\|^2_2 \le 1 \quad \forall k \in  \{1,\dots,K \},
	\end{aligned}
	\label{problem_csc}
	\end{equation}
	where $\bb$ are input images, $\bx=[\bx_1^T,\dots,\bx_K^T]^T$ is with sparsity constraint and $\bA=[\bA_1,\dots,\bA_K]$ is a concatenation of Toeplitz matrices representing convolution with the respective filters $\ba_k$. $\lambda$ is the sparsity parameter and $K$ is the number of the kernels. According to the framework~\cite{gu2015convolutional}, we introduce a diagonal or block-diagonal matrix $\bM$, and then reformulate Eq.~\ref{problem_csc} as
	\begin{equation}
	\small
	\begin{aligned}
	&\mathop {\arg \min }\limits_{\bA,\bx} f_1(\bA\bx)+\sum_{k=1}^K(f_2(\bx_k)+f_3(\ba_k)),
	\end{aligned}
	\label{problem_csc_sum}
	\end{equation}
	
	\noindent where 
	
	\begin{equation}
	\small
	\begin{aligned}
	f_1(\bv)&=\frac{1}{2}\left\|\bb-\bM\bv\right\|_F^2,\\
	f_2(\bv)&=\lambda\left\|\bv\right\|_1,\\ 
	f_3(\bv)&=ind_c(\bv).
	\end{aligned}
	\label{eq:proximal_csc}
	\end{equation}
	
	\noindent Here, $ind_c(\cdot)$ is an indicator function defined on the convex set of the constraints $C = \{\mathbf{x} \vert  \left\| \mathbf{Sx} \right\|_2^2 \le 1 \}$. The solution to Eq.~\ref{problem_csc_sum} by ADMM with proximal operators is detailed in~\cite{parikh2014proximal,heide2015fast}. However, it suffers from asynchronous convergence, which leads to a sub-optimal solution. Instead, we introduce our CoGD for this optimization. Note that in this case, similar to Eq.~\ref{bt_x}, we have
	\begin{equation}
	\small
	\begin{aligned}
	&\ \ \ \ \ \ \hat{\bx}_k =
	\begin{cases}
	P({\bx}^{t+1}_k,{\bx}^{t}_k)& if\ (\neg s(\bx_k))\wedge(s(\bA_{k}))=1\\
	%P(x^{T,t+1},x^{T,t})& otherwise.
	{\bx}^{t+1}_k& otherwise
	\end{cases}\\
	\end{aligned}
	\label{bt_csc}
	\end{equation}
	%	where $ \hat{G}(\bA,\bx) = (\bA\bx^{t}-\bb)^T,\ \frac{\partial \bA_k}{\partial {\bx_k}}=\mathbf{1}$, and  $\mathbf{1}$ denotes an vector of all ones. $\alpha_{\bA}$ can be calculated by the sorted result of $\{\bA_1,\dots,\bA_K\}$. $\alpha_{\bx}$ is set as a constant. More details can refer to the experiments. Finally, to jointly solve $\bA$ and $\bx$, we follow the proposed approach in sec.~\ref{sec:theory} to solve the two coupled variables iteratively, yielding Alg.~\ref{alg_csc}.
	where $ \hat{G}(\bA,\bx) = (\bA\bx^{t}-\bb)^T$. $\beta$, $\alpha_{\bA}$ and  $\alpha_{\bx}$ are detailed in  the experiments. Finally, to jointly solve $\bA$ and $\bx$, we follow the proposed approach in sec.~\ref{sec:theory} to solve the two coupled variables iteratively, yielding Alg.~\ref{alg_csc}.

	\input{algorithm/algorithm_CSC.tex}

	\subsection{CoGD for CNNs Pruning}
	\subsubsection{CNNs Pruning based on a Bilinear Model}
	Channel pruning has received increased attention recently for compressing convolutional neural networks (CNNs). Early works in this area tended to directly prune the kernel based on simple criteria like the norm of kernel weights~\cite{li2016pruning} or a greedy algorithm~\cite{luo2017thinet}.
	More recent approaches have sought to formulate network pruning as a bilinear optimization problem with soft masks and sparsity regularization~\cite{he2017channel,ye2018rethinking,Huang2017Data,Lin2019Towards}.
	
	Based on the framework of~\cite{he2017channel,ye2018rethinking,Huang2017Data,Lin2019Towards}, we apply our CoGD for channel pruning.
	%We construct a generative adversarial network framework by following ~\cite{Lin2019Towards}, which consist of three sub-networks. The first part is the pre-trained network. We regard the pre-trained network as baseline and the output of pre-trained network as the real label to replace the dataset label. The second part is the pruned network, which is set as a generator. The pruned network has the same network architecture and shares the same initial network weights as the pre-trained network. The final part is the discriminator network. We use the discriminator to distinguish the output distribution between the pre-trained network and pruned network.
	In order to prune the channel of the pruned network, the soft mask $\bm$ is introduced after the convolutional layer to guide the output channel pruning and has a bilinear form of 
	\begin{equation}
	F_j^{l + 1} =  f(\sum\limits_i {F_i^l \otimes (W_{{{i,j}}}^l}{\bm_j}) ),
	\label{eq_network_pruning}
	\end{equation}
	where \(F_j^{l}\) and \(F_j^{l + 1}\) are the \( i \)-th input  and the \( j \)-th output feature maps at the \( l \)-th and \( l+1 \)-th layer. \( W^l_{i,j}\) are convolutional filters which  correspond to the soft mask \(\bm\). \( \otimes \) and \( f( \cdot )\) refer to convolutional operator and activation,  respectively.  
	
	\input{algorithm/algorithm_pruning.tex}
		
	In this framework, the soft mask $\bm$ can be learned end-to-end in the back propagation process.
	\zhuo{To be consistent with other pruning works, we use $W$ and $\bm$ instead of $\bA$ and $\bx$ in this part. }
	Then a general optimization function for network pruning with soft mask is formulated with a bilinear form as
	\begin{equation}
	\mathop {\arg \min }\limits_{W,\bm}
	\mathcal{L} ( W{\bm}) +\lambda{\left\| \bm \right\|_1}+R({W}),
	\label{problem_pruning}
	\end{equation}
	where $\mathcal{L} ( W{\bm})$ is the loss function, which will be detailed in following.  With the sparsity constraint on $\bm$, the convolutional filters with zero value in the corresponding soft mask are regarded as useless filters, which means that these filters and their corresponding channels in the feature maps have no significant  contribution to the subsequent computation and will ultimately be removed. However, there is a dilemma in the  pruning-aware training in that the pruned filters are not evaluated well before they are pruned, which leads to sub-optimal pruning. Particularly,  $\bm$ and the corresponding kernels do not become sparse in a synchronous manner, which will cause an inefficient training problem for CNNs, which is allured in our paper.  To address the  problem, we again apply our CoGD to calculate the soft mask, by reformulating Eq.~\ref{bt_x} as 
	\begin{equation}
	\small
	\begin{aligned}
	\hat{\bm}^{l,t+1}_j\!=&
	\begin{cases}
	P({\bm_j}^{l,t+1},{\bm_j}^{l,t})&if\ (\neg s(\bm_j^{l,t}))\wedge s(\sum_i W_{i,j}^l)\!=\!1\\
	%P(x^{T,t+1},x^{T,t})& otherwise.
	{\bm}^{l,t+1}_j & otherwise,
	\end{cases}
	\end{aligned}
	\label{bt_pruning}
	\end{equation}
	where $W_{i,j}$ represents the 2D kernel of the  $i$-th input channel of the $j$-th filter. 
		%The threshold $\alpha$ is associated with the proportion of $\bm$ that will be particularly updated by our CoGD. 
	$\beta$,	$\alpha_{W}$  and $\alpha_{\bm}$ are detailed in our experiments.
	\zhuonew{To further investigate the form of $\hat{G}$ for CNNs pruning, we have	
	\begin{equation}
	\begin{aligned}
	\frac{\partial \mathcal{L}}{\partial W_{i,j,p,q}}&=\bm_j \sum_h^H \sum_w^W ((\frac{\partial \mathcal{L}}{\partial F_j^{l+1}})_{h,w}F^l_{i,h+p-1,w+q-1})\\
	&=\bm_j\hat{g}_{j,i},	
	\end{aligned}
	\label{eq_pg}
	\end{equation}
	where the size of input feature maps $F^l_i$ is $H\times W$. $p$ and $q$ is the kernel indexs for convolutional filters $W^l_{i,j}$. Noted that the above partial gradient has usually been calculated by the autograd package in common deep learning frameworks, \eg, Pytorch \cite{paszke2019pytorch}. Therefore, based on Eq.~\ref{eq_pg}, we simplify the calculation of $\hat{G}$ as following:
	\begin{equation}
	\hat{G} = \frac{\partial \mathcal{L}}{\partial W_{i,j}}/\bm_j.
	\label{eq_hatg}
	\end{equation}
    }
    %$\hat G =   \frac{\partial \mathcal{L}}{\partial W_j}/\bm_j$  is detailed in the supplemented file.
	
	%\subsubsection{CNN Pruning based on CoGD and GAL}
	
	We prune CNNs based on the new mask $\hat{\bm}$ in Eq.~\ref{bt_pruning}. We use GAL\footnote{We implement our method based on their source code.} \cite{Lin2019Towards}  as an example to describe our CoGD for CNNs pruning. A pruned network obtained through GAL, with \(\ell_{1}\)-regularization on the soft mask is used to approximate the pre-trained network by aligning their outputs. The discriminator \(D\) with weights \(W_D\) is introduced to discriminate between the output of pre-trained network and pruned network, and the pruned network generator  \(G\) with weights \(W_G\) and soft mask $\bm$ is learned together with \(D\) by using the knowledge from supervised features of baseline.
	%We construct a generative adversarial network framework by following ~\cite{Lin2019Towards}, which consist of three sub-networks. The first part is the pre-trained network $W_B$. We regard the pre-trained network as baseline and the output of pre-trained network as the real label to replace the dataset label. The second part is the pruned network $W_G$, which is set as a generator. The pruned network has the same network architecture and shares the same initial network weights as the pre-trained network. The final part is the discriminator network $W_G$. We use the discriminator to distinguish the output distribution between the pre-trained network and pruned network.	
	
	Accordingly, \(\bm\), \(\hat{\bm}\), the pruned network weights \(W_G\) and the discriminator weights \(W_D\) are all learned by solving the optimization problem as follows 
	\begin{equation}
	\small
	\begin{aligned}
	\arg \mathop {\min }\limits_{{W_G},\bm} \mathop {\max }\limits_{{W_D},\hat{\bm}} &{\mathcal{L}_{Adv}}({W_G},\hat{\bm},{W_D})+ {\mathcal{L}_{data}}({W_G},\hat{\bm})\\
	+ &{\mathcal{L}_{reg}}({W_G},\bm,{W_D}).
	\end{aligned}
	\label{optimization}
	\end{equation}
	where $\mathcal{L} ( W{\bm}) = {\mathcal{L}_{Adv}}({W_G},\hat{\bm},{W_D})+ {\mathcal{L}_{data}}({W_G},\hat{\bm})$ and 	 ${\mathcal{L}_{reg}}({W_G},\bm,{W_D})$ is related to $\lambda{\left\| \bm \right\|_1}+R({W})$ in Eq. \ref{problem_pruning}.  \({\mathcal{L}_{Adv}}({W_G},\hat{\bm},{W_D})\) is the adversarial loss to train the two-player game between the pre-trained network and the pruned network that compete with each other. %Note that we did not introduce additional losses to  \cite{Lin2019Towards}. 
	More details of the algorithm are shown in Alg.~\ref{alg_gal}.

	\begin{figure}[!t]
		\centering
		\includegraphics[scale=0.40]{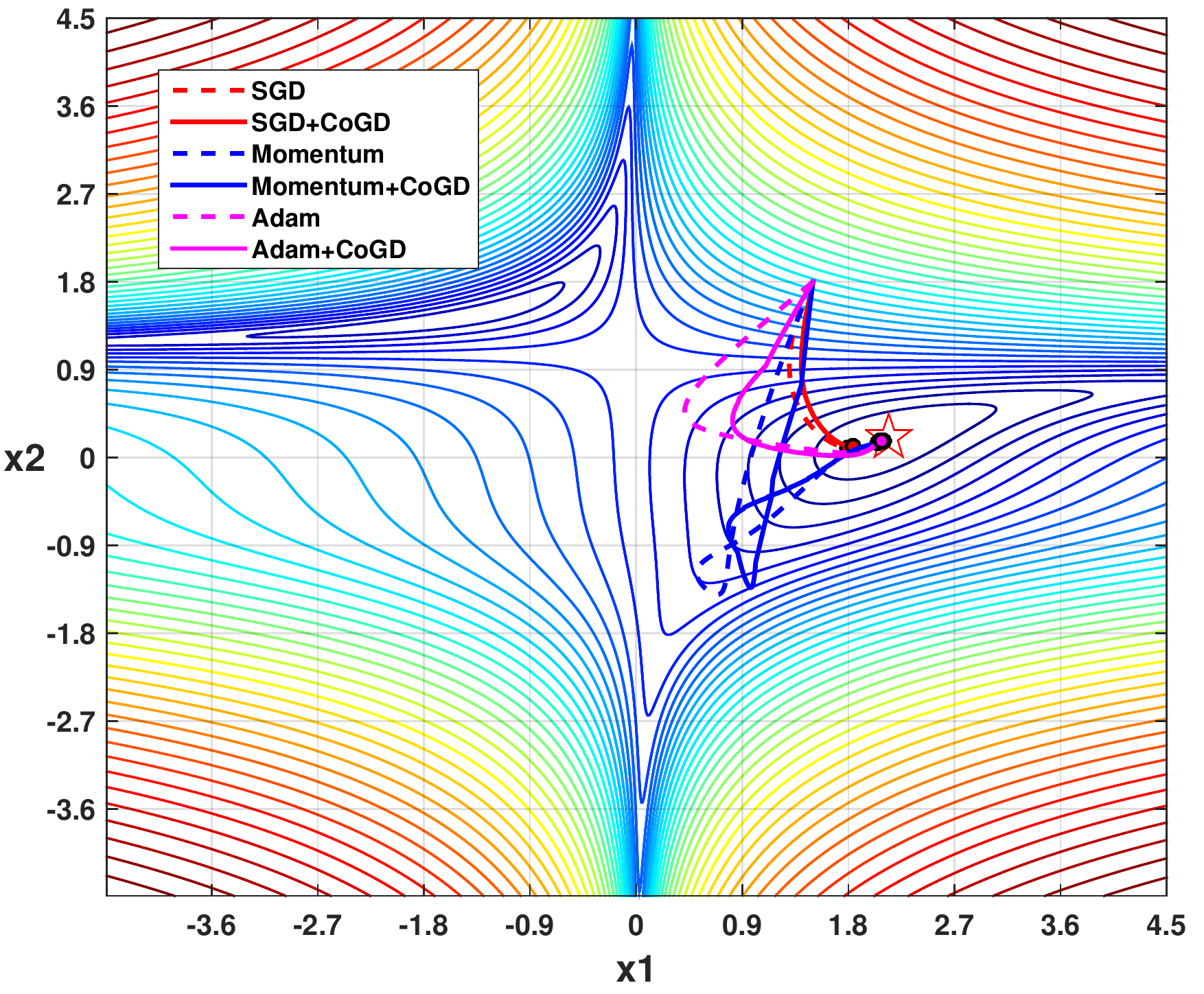}%
		\caption{\zhuo{The contour map of beale functions. It can be seen that algorithms with CoGD have a short optimization path compared with their counterparts, which shows that CoGD facilitates efficient and sufficient training.}}
		\label{optimizer}
		\vspace{-0.4cm}
	\end{figure}
	
	The  advantages of CoGD in network pruning are three-fold. First, our method optimizes the bilinear pruning model, which leads to a synchronous gradient convergence. Second, the process is controllable by the threshold, which makes the pruning rate easy to adjust. Third, our method for pruning CNNs is generic and can be built atop of other state-of-the-art networks such as~\cite{he2017channel,ye2018rethinking,Huang2017Data,Lin2019Towards}, for better performance. 
	
	\section{Experiments}

	\subsection{CoGD for Toy problem}
	
	\zhuo{We first use a toy problem as an example to illustrate the superiority of our algorithm based on  the optimization path. 
	In Fig.~\ref{optimizer}, we implement CoGD algorithm on three widely used optimization methods, \ie, `SGD', `Momentum' and `Adam'. We solve the beale function \footnote{$beate(x_1,x_2)=(1.5-x_1+x_1x_2)^2+(2.25-x_1+x_1x_2^2)^2+(2.62-x_1+x_1x_2^3)^2$.} with additional constraint $F(x_1,x_2)=beale(x_1,x_2)+\left\|x_1\right\|+x_2^2$. This function has the same form as Eq.~\ref{eq:bl_sparse} and can be  regraded as a bilinear problem with the  part $x_1  x_2$. The learning rate $\eta_2$ is set as $0.001$, $0.005$, $0.1$ for `SGD', `Momentum' and `Adam' respectively. The threshold $\alpha_{x_1}$ and $\alpha_{x_2}$  for CoGD are set as $1$, $0.5$. \zhuo{$\beta = 0.001 \eta_2 \bc^{t}$} with $\frac{\partial x_2}{\partial {x_1}} =  \frac{\Delta x_2}{\Delta x_1}$, where $\Delta$ denotes the difference of variable over the epoch. $\frac{\Delta x_2}{\Delta x_1} =\mathbf{1}$, when $\Delta x_2$ or $x_2$  approaches to be zero. $0.1$ in $\beta$ is used to enhance the gradient  of  $x_2$.
	 The total  iterations is $200$.  It can be seen that algorithms with CoGD have short optimization paths compared with their counterparts, which show that CoGD facilitates an efficient and sufficient training.}
	
%	\eta_2 \bc^{t}

	\begin{figure}[!t]
		\centering
		\subfloat{
			\centering
			\label{filters1}
			\includegraphics[width=0.48\textwidth]{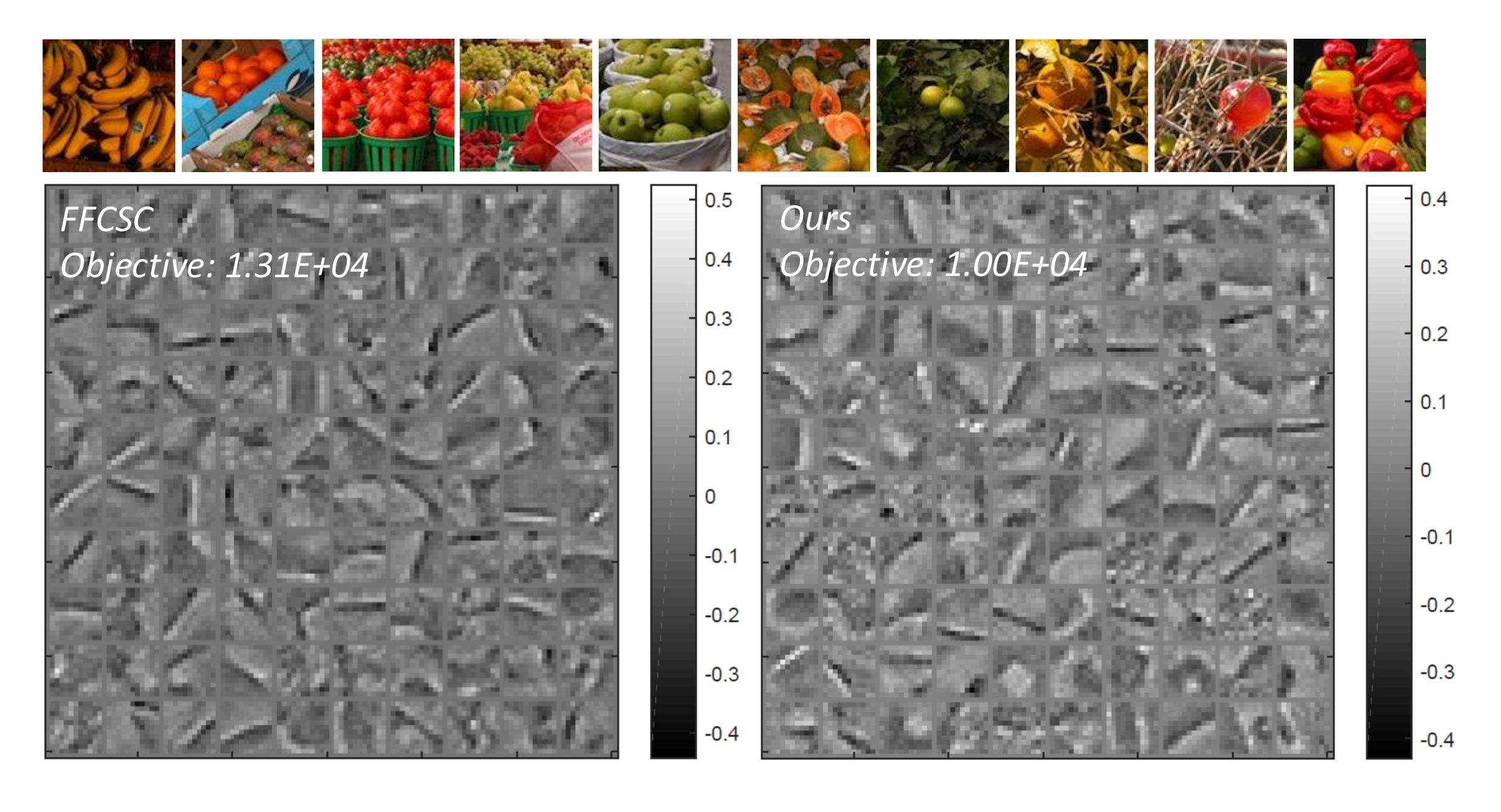}}
		
		\subfloat{
			\centering
			\label{filters2}
			\includegraphics[width=0.48\textwidth]{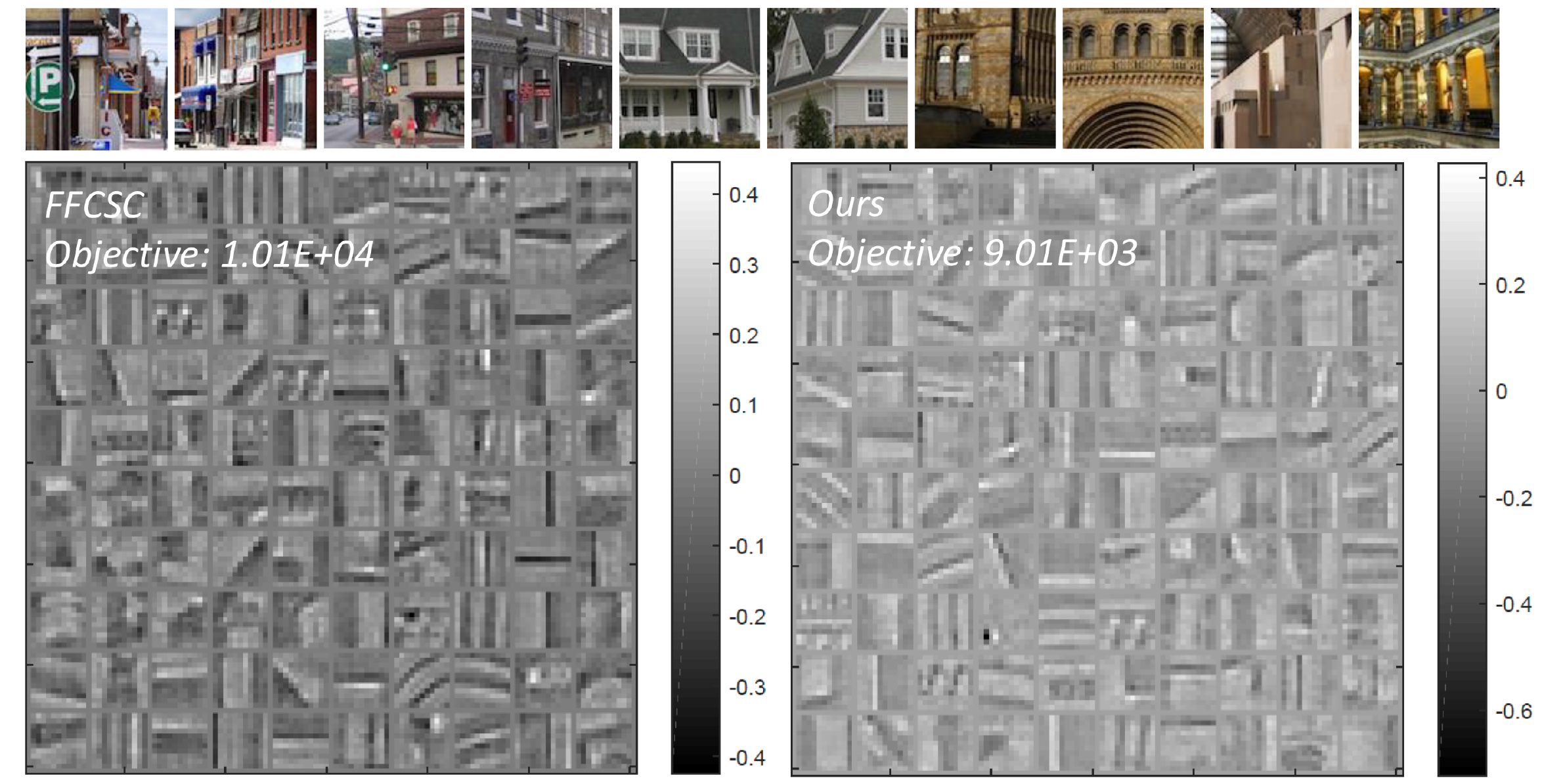}}
		\caption{Filters learned on fruit and city datasets. Thumbnails of the datasets along with filters learned  with FFCSC~\cite{heide2015fast} (left) and with our proposed method (right) are shown. Note that our method achieves lower objective. }
		\label{fig:filters}
		\vspace{-0.4cm}
	\end{figure}

	\subsection{Convolutional Sparse Coding}
	
	\textbf{Datasets.} We evaluate our method on two  publicly available datasets: the fruit dataset~\cite{zeiler2010deconvolutional} and the city dataset~\cite{zeiler2010deconvolutional}. They are commonly used to evaluate CSC methods~\cite{zeiler2010deconvolutional,heide2015fast} and each of them consists of ten images with 100 $\times$ 100 resolution. 
	
	To evaluate the quality of the reconstructed images, we use two common metrics, the peak signal-to-noise ratio (PSNR, unit:dB) and the structural similarity (SSIM). The higher PSNR and SSIM are, the better visual quality is achieved for the reconstructed image.
	
	\textbf{Implementation Details.} Our model is implemented based on~\cite{gu2015convolutional}, by only using the projection function to achieve \zhuo{a better convergence}. We use 100 filters with size 11$\times$11. \zhuo{$\alpha_{\bx}$ is set as the mean of $\left\|\bx_k\right\|_1$. $\alpha_{\bA}$ is calculated by the median of the sorted result of $\bA_k$.  } For a fair comparison, we use the same hyperparameters ($\eta_2$) in both our method and~\cite{gu2015convolutional}. Similarly, $\beta = 0.1 \eta_2 \bc^{t}$ with $\frac{\partial \bA}{\partial {x_j}} =  \frac{\Delta \bA}{\Delta x_j}$. $0.1$ in $\beta$ is used to enhance the gradient  of $A$.
	
	\begin{table*}[!t]
		\centering
		\small
		\caption{Reconstruction results for filters learned with our proposed method and with FFCSC~\cite{heide2015fast}. With the exception of 6 images, our method results in better PSNR and SSIM.}
		\scalebox{0.83}{
			\begin{tabular}{c c c c c c c c c c c c c c}
				\hline\hline
				Dataset&Fruit& 1 & 2 & 3 & 4 & 5 & 6 & 7 & 8 & 9 & 10 & Average \\ \hline
				\multirow{2}{*}{PSNR (dB)}&~\cite{heide2015fast} & 30.90 & \textbf{29.52} & 26.90 & 28.09 & 22.25 & \textbf{27.93} & \textbf{27.10}& 27.05 & 23.65 & 23.65 & 26.70 \\ 
				&CoGD  & \textbf{31.46} & 29.12 & \textbf{27.26} & \textbf{28.80} & \textbf{25.21} & 27.35 & 26.25 & \textbf{27.48} & \textbf{25.30} & \textbf{27.84} & \textbf{27.60} \\ \hline	
				\multirow{2}{*}{SSIM}& ~\cite{heide2015fast} & 0.9706 & \textbf{0.9651} & 0.9625 & \textbf{0.9629} & 0.9433 & 0.9712 & 0.9581 & 0.9524 & 0.9608 & 0.9546 & 0.9602 \\ 
				&CoGD  & \textbf{0.9731} & 0.9648 & \textbf{0.9640} & 0.9607 & \textbf{0.9566} & \textbf{0.9717} & \textbf{0.9587} & \textbf{0.9562} & \textbf{0.9642} & \textbf{0.9651} & \textbf{0.9635} \\ \hline\hline		
				
				Dataset&City& 1 & 2 & 3 & 4 & 5 & 6 & 7 & 8 & 9 & 10 & Average \\ \hline
				\multirow{2}{*}{PSNR (dB)}&~\cite{heide2015fast} & 30.11 & 27.86 & \textbf{28.91} & \textbf{26.70} & 27.85 & 28.62 & 18.63 & 28.14 & 27.20 & 25.81 & 26.98\\ 
				&CoGD  & \textbf{30.29}& \textbf{28.77}& 28.51& 26.29& \textbf{28.50}& \textbf{30.36}& \textbf{21.22}& \textbf{29.07}& \textbf{27.45}& \textbf{30.54}& \textbf{28.10}  \\ \hline
				\multirow{2}{*}{SSIM}&~\cite{heide2015fast} & 0.9704 & \textbf{0.9660} & \textbf{0.9703} & 0.9624 & 0.9619 & 0.9613 & 0.9459 & 0.9647 & 0.9531 & 0.9616 & 0.9618 \\ 
				&CoGD  & \textbf{0.9717}& \textbf{0.9660}& 0.9702& \textbf{0.9628}& \textbf{0.9627}& \textbf{0.9624}& \textbf{0.9593}& \textbf{0.9663}& \textbf{0.9571}& \textbf{0.9632}& \textbf{0.9642}  \\ \hline\hline
		\end{tabular}}
		\label{tab:reconstruction}
		\vspace{-0.2cm}
	\end{table*}
	
	\begin{table*}[!t]
		\centering
		\small
		\caption{Inpainting results for filters learned with our proposed method and with FFCSC~\cite{heide2015fast}. All reconstructions are performed for 75\% subsampling. Our method achieves better PSNR and SSIM in all cases.}
		\scalebox{0.83}{
			\begin{tabular}{c c c c c c c c c c c c c c}
				\hline\hline
				Dataset&Fruit& 1 & 2 & 3 & 4 & 5 & 6 & 7 & 8 & 9 & 10 & Average \\ \hline
				\multirow{2}{*}{PSNR (dB)}&~\cite{heide2015fast} & 25.37 & 24.31 & 25.08 & 24.27 & 23.09 & 25.51 & 22.74& 24.10 & 19.47 & 22.58 & 23.65 \\ 
				&CoGD  & \textbf{26.37} & \textbf{24.45} & \textbf{25.19} & \textbf{25.43} & \textbf{24.91} & \textbf{27.90} & \textbf{24.26} & \textbf{25.40} & \textbf{24.70} & \textbf{24.46} & \textbf{25.31} \\ \hline	
				\multirow{2}{*}{SSIM}&~\cite{heide2015fast} & 0.9118 & 0.9036 & 0.9043 & 0.8975 & 0.8883 & 0.9242 & 0.8921 & 0.8899 & 0.8909 & 0.8974 & 0.9000 \\ 
				&CoGD  & \textbf{0.9452} & \textbf{0.9217} & \textbf{0.9348} & \textbf{0.9114} & \textbf{0.9036} & \textbf{0.9483} & \textbf{0.9109} & \textbf{0.9041} & \textbf{0.9215} & \textbf{0.9097} & \textbf{0.9211} \\ \hline\hline		
				
				Dataset&City& 1 & 2 & 3 & 4 & 5 & 6 & 7 & 8 & 9 & 10 & Average \\ \hline
				\multirow{2}{*}{PSNR (dB)}&~\cite{heide2015fast} & 26.55 & 24.48 & 25.45 & 21.82 & 24.29 & 25.65 & 19.11 & 25.52 & 22.67 & 27.51 & 24.31\\ 
				&CoGD  & \textbf{26.58}& \textbf{25.75}& \textbf{26.36} & \textbf{25.06} & \textbf{26.57}& \textbf{24.55}& \textbf{21.45}& \textbf{26.13}& \textbf{24.71}& \textbf{28.66}& \textbf{25.58}  \\ \hline
				\multirow{2}{*}{SSIM}&~\cite{heide2015fast} & 0.9284 & 0.9204 & 0.9368 & 0.9056 & 0.9193 & 0.9202 & 0.9140 & 0.9258 & 0.9027 & 0.9261 & 0.9199 \\ 
				&CoGD  & \textbf{0.9397}& \textbf{0.9269}& \textbf{0.9433}& \textbf{0.9289}& \textbf{0.9350}& \textbf{0.9217}& \textbf{0.9411}& \textbf{0.9298}& \textbf{0.9111}& \textbf{0.9365}& \textbf{0.9314}  \\ \hline\hline
		\end{tabular}}
		\label{tab:inpainting}
		\vspace{-0.4cm}
	\end{table*}	
	
	\begin{figure}[!t]
		\centering
		\includegraphics[width=0.45\textwidth]{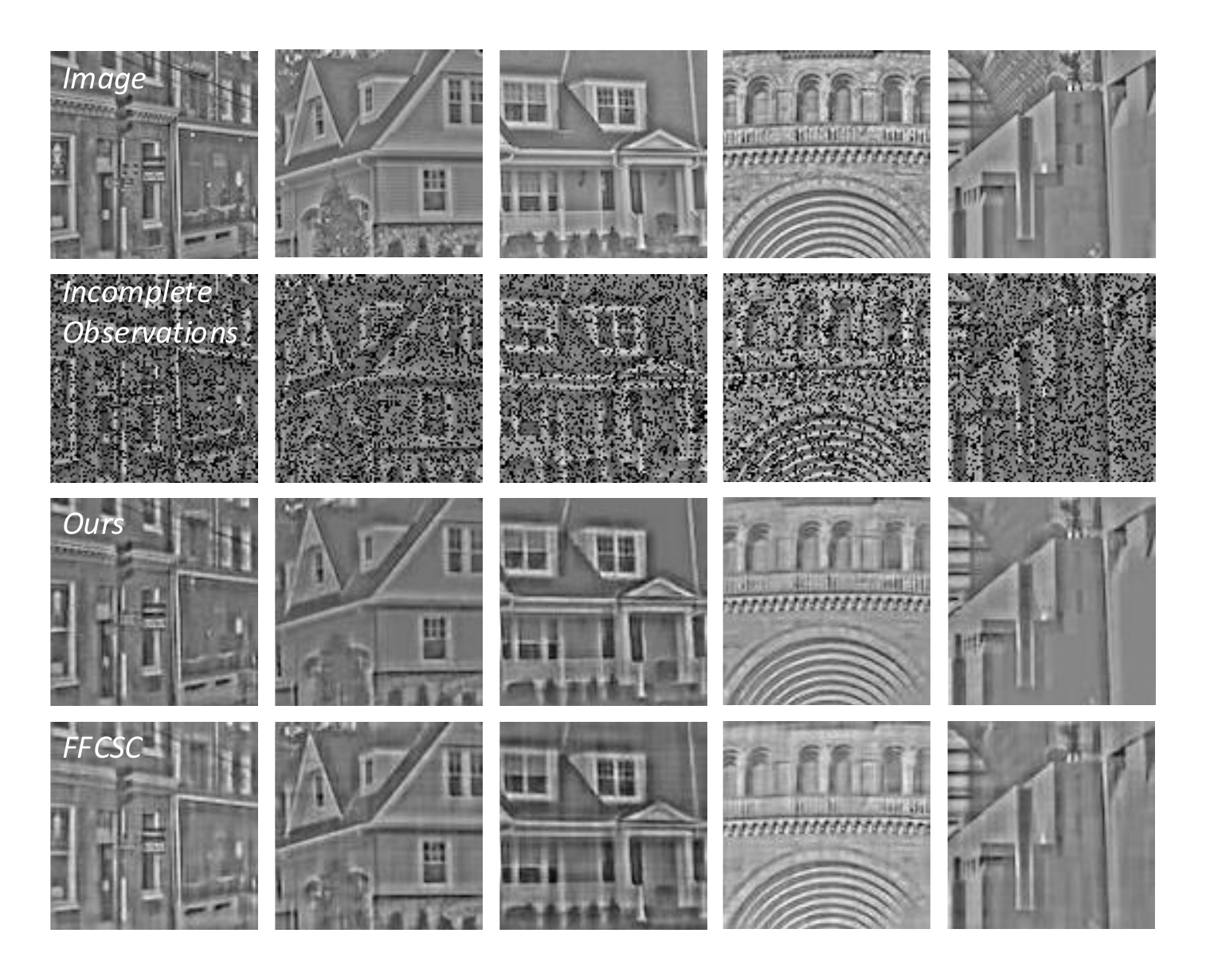}
		\caption{Inpainting for normalized city dataset. From top to bottom: original images, incomplete observations, reconstructions with FFCSC~\cite{heide2015fast}, reconstructions with our proposed algorithm. }
		\label{Inpainting}
		\vspace{-0.4cm}
	\end{figure}

	\textbf{Results.} We evaluate CSC with our algorithm for two tasks, including image reconstruction and image inpainting.
	
	For \textbf{image reconstruction}, we reconstruct the images on fruit and city datasets respectively. We train 100 11$\times$11- sized filters, and compare these filters with  FFCSC~\cite{heide2015fast}. Fig.~\ref{fig:filters} shows the resulting filters after convergence within the same
	%- our 20 iterations v.s. \cite{heide2015fast}'s 
	20 iterations. Comparing our method to FFCSC, we observe that our method converges with a lower loss. Furthermore, we compare PSNR and SSIM of our method with FFCSC in Tab.~\ref{tab:reconstruction}. In most cases, our method achieves better PSNR and SSIM. The average PSNR and SSIM improvements are 1.01 db and 0.003.
	
	For \textbf{image inpainting}, we randomly sample the data with a 75\% subsampling rate, to obtain the incomplete data. Similar to~\cite{heide2015fast}, we test our method on contrast-normalized images. We first learn filters from all the incomplete data with the help of $\bM$, and then reconstruct that incomplete data by fixing the learned filters.
	We show some inpainting results of normalized data in Fig.~\ref{Inpainting}. Moreover, to compare with FFCSC, inpainting results on the fruit and city datasets are shown in Tab.~\ref{tab:inpainting}. Note that our method achieves better PSNR and SSIM in all cases. The average PSNR and SSIM improvements are an  impressive 1.47 db and 0.016.

	\subsection{Network Pruning}

	We have evaluated CoGD algorithm on network pruning using the CIFAR-10 and ILSVRC12 ImageNet datasets for classification. ResNets and  MobileNetV2 are the backbone networks to build our framework.
	%\zhuonew{ We note that CoGD-$a$ means that the projection proportion is $a$\% with an approximate pruning rate of $1-a$\%, which is associated with $\alpha_W$.}
	
	\begin{table}[!t]
		\small
		\caption{Pruning results of ResNet-18/110 and MobilenetV2 on CIFAR-10. M means million (\({10^6}\)).}
		\centering
		\scalebox{0.83}{
			\begin{tabular}{c c c c}
				\hline\hline
				Model     &FLOPs (M) & Reduction & Accuracy/+FT (\%)  \\ \hline 
				ResNet-18\cite{He2016DeepNew}  &555.42 & - &95.31    \\ 
				CoGD-0.5 &274.74 & 0.51$\times$ &95.11/95.30\\ 
				%CoGD-0.75 &423.87 & 0.24$\times$ &\textbf{95.41}/-\\ 
				%CoGD-0.95 &525.46 & 0.05$\times$ &\textbf{95.49}/-\\ 
				\hline							
				ResNet-110\cite{He2016DeepNew} &252.89 & - &93.68      \\
				GAL-0.1\cite{Lin2019Towards} &205.70 & 0.20$\times$ &92.65/93.59  \\ 
				GAL-0.5\cite{Lin2019Towards} &130.20 & 0.49$\times$ &92.65/92.74  \\ 
				CoGD-0.5 &  95.03  & 0.62$\times$ & 93.31/93.45                  \\ 
				%CoGD-0.75 &  125.35 & 0.50$\times$ & 93.50/\textbf{93.81}         \\ 
				CoGD-0.8 &  135.76 & 0.46$\times$ & 93.42/93.66         \\ 
				\hline	
				MobileNet-V2\cite{Sandler2018MobileNetV2} &91.15 & - & 94.43 \\ 
				CoGD-0.5 &50.10 & 0.45$\times$ &94.25/- \\ 
				%CoGD-0.75 &67.82 & 0.26$\times$ &\textbf{94.58}/-\\ 
				%CoGD-0.95 &84.37 & 0.07$\times$ &\textbf{94.83}/-\\ 
				\hline\hline					
		\end{tabular}}
		\label{cifar10_res}
		\vspace{-0.5cm}
	\end{table}

	\subsubsection{Datasets and Implementation Details}
	
	\textbf{Datasets: }CIFAR-10 is a natural image classification dataset containing a training set of $50,000$ and a testing set of $10,000$ \(32 \times 32\)  color images distributed over 10 classes including: airplanes, automobiles, birds, cats, deer, dogs, frogs, horses, ships, and trucks.
	
	The ImageNet object classification dataset is more challenging due to its large scale and greater diversity. There are $1,000$ classes, $1.2$ million training images and 50k validation images.

	\textbf{Implementation Details.}
	We use PyTorch to implement our method with $3$ NVIDIA TITAN V GPUs. The weight decay and the momentum are set to $0.0002$  and  $0.9$ respectively. The hyper-parameter \( \lambda_\bm\) is selected through cross validation in the range $[0.01, 0.1]$ for ResNet and MobileNetv2. The drop rate is set to $0.1$. The other training parameters are described on a per experiment basis.
	
	\zhuo{To better demonstrate our method, we denote CoGD-a as  an approximated pruning rate of $(1-a)$\% for coresponding channels.  $a$  is associated with the threshold $\alpha_W$, which is given  by its sorted result. For example, if $a = 0.5$,  $\alpha_W$ is the median of sorted result. $\alpha_{\bm}$ is set to be $0.5$ for easy implementation.} 
	%Being different from toy problem and CSC, 
	Similarly, $\beta = 0.001 \eta_2 \bc^{t}$ with $\frac{\partial \bW}{\partial {m_j}} =  \frac{\Delta \bW}{\Delta m_j}$.
	%the gradient calculation in CNNs becomes more complex, we set $\beta = 1$. We also find that a value between $1$ and $1.5$ affects little on the final performance. 
	Note that our training cost is similar to \cite{Lin2019Towards}, since we use our method once per epoch without extra cost.

	%	\subsection{ Comparison with State-of-the-Arts}
	\subsubsection{Experiments on CIFAR-10}
	We evaluated our method on CIFAR-10 for two popular networks, ResNets and MobileNetV2. The stage kernels are set to 64-128-256-512 for ResNet-18  and 16-32-64  for ResNet-110.
	For all networks, we add the soft mask only after the first convolutional layer within each block to prune the output channel of the current convolutional layer and input channel of next convolutional layer, simultaneously.	The mini-batch size is set to be 
	$128$ for $100$ epochs, and the initial learning rate is set to $0.01$,
	scaled by $0.1$ over $30$ epochs.% The $\frac{\partial \bW_j}{\partial \bm_j}$ is set to $\mathbf{1}$. %\zhuonew{$\alpha_{\bm}$ is set as $0.5$. $\alpha_W$ is calculated based by sorted result,  which is determined by $a$. For example, if $a$ is equal to $0.5$,  $\alpha_W$ is obtained by the median of sorted result. Note that our training cost is similar to the work~\cite{Lin2019Towards}, since we use our method once per epoch without extra cost.}
	
	\textbf{Fine-tuning.} For fine-tuning, we only reserve the student model. According to the ‘zero’s in each soft mask, we remove the corresponding output channels of current convolutional layer and corresponding input channels of next convolutional layer. We then obtain a pruned network with fewer parameters and FLOPs. 
	We use the same batch size of $256$ for $60$ epochs as in training. The initial learning rate is changed to be $0.1$ and scaled by $0.1$ over $15$ epochs. Note that a similar fine-tuning strategy was used in GAL.
	
	%\zhuonew{\textbf{Convergence.} }

	\textbf{Results.} Two kinds of networks are tested on the CIFAR-10 database - ResNets and MobileNet-V2.
	
	For \textbf{ResNets}, as is shown in Tab.~\ref{cifar10_res}, our method achieves comparable results.
	Compared to the pre-trained network for ResNet-18 with $95.31$\% accuracy, CoGD-$0.5$ achieves a \({\rm{0.51}} \times \) FLOPs reduction with only a \(0.01\% \) drop in accuracy. %When fewer channels are pruned, CoGD performs  better. CoGD-$0.75$  has a \({\rm{0.24}} \times \) FLOPs reduction with $0.10$\% improvement as the pre-trained network. 
	%CoGD-$0.95$ achieves the best performance $95.49$\% but with a \({\rm{0.05}} \times \) FLOPs reduction. 
	Among other structured pruning methods for ResNet-110, %CoGD achieves a smaller accuracy drop than GAL-$0.5$ ($93.81$\% v.s. $92.74$\%) with a similar FLOPS reduction.  
	CoGD-$0.5$ has a larger FLOPs reduction than GAL-$0.1$ ($95.03M$ v.s. $205.70M$), but with the similar accuracy ($93.45$\% v.s. $93.59$\%). These results demonstrate that our method is able to prune the network efficiently and generate a more compressed model with higher performance.
	
	\begin{table}[!t]
		\caption{\zhuo{Pruning results of ResNet-50 on ImageNet. B means billion (\({10^9}\)).}}
		\centering
		\scalebox{0.83}{
			\begin{tabular}{c c c c}
				\hline\hline
				Model      & FLOPs (B) & Reduction & Accuracy/+FT (\%)  \\ \hline
				ResNet-50\cite{He2016DeepNew}  & 4.09 & - & 76.24   \\ 
				ThiNet-50\cite{luo2017thinet}  & 1.71 & 0.58$\times$ & 71.01               \\ 
				ThiNet-30\cite{luo2017thinet}  & 1.10 & 0.73$\times$ & 68.42               \\ 
				CP\cite{he2017channel}         & 2.73 & 0.33$\times$ & 72.30               \\ 
				GDP-0.5\cite{lin2018accelerating}  & 1.57 & 0.62$\times$ & 69.58               \\ 
				GDP-0.6\cite{lin2018accelerating}  & 1.88 & 0.54$\times$ & 71.19               \\ 
				SSS-26\cite{huang2018data}     & 2.33 & 0.43$\times$ & 71.82                \\ 
				SSS-32\cite{huang2018data}     & 2.82 & 0.31$\times$ & 74.18                \\ 
				RBP\cite{zhou2019accelerate}   & 1.78 & 0.56$\times$ & 71.50                \\ 
				RRBP\cite{zhou2019accelerate}  & 1.86 & 0.55$\times$ & 73.00                \\ 	
				GAL-0.1\cite{Lin2019Towards}   & 2.33 & 0.43$\times$ & -/71.95              \\ 
				GAL-0.5\cite{Lin2019Towards}   & 1.58 & 0.61$\times$ & -/69.88              \\ 
				%CoGD-0.35 &  -    & &  -/-         \\ 
				%CoGD-0.3 &  -    & &  -/-         \\ 
				%CoGD-0.75 &  -    & &  -/-         \\ 
				CoGD-0.5 &  2.67    & 0.35$\times$ &  75.15/75.62         \\ 
				\hline\hline
		\end{tabular}}
		\label{imagenet_res50}
		\vspace{-0.3cm}
	\end{table}	

%	\begin{table}[!t]
%		\caption{\zhuonew{Pruning results of ResNet-50 on ImageNet. B means billion (\({10^9}\)).}}
%		\centering
%		\scalebox{0.83}{
%			\begin{tabular}{c c c}
%				\hline\hline
%				Model      & FLOPs (B) & Accuracy/+FT (\%)  \\ \hline
%				ResNet-50\cite{He2016DeepNew}  & 4.09 & 76.24   \\ 
%				ThiNet-50\cite{luo2017thinet}  & 1.71 & 71.01               \\ 
%				ThiNet-30\cite{luo2017thinet}  & 1.10 & 68.42               \\ 
%				CP\cite{he2017channel}         & 2.73 & 72.30               \\ 
%				GDP-0.5\cite{lin2018accelerating}  & 1.57 & 69.58               \\ 
%				GDP-0.6\cite{lin2018accelerating}  & 1.88 & 71.19               \\ 
%				SSS-26\cite{huang2018data}     & 2.33 & 71.82                \\ 
%				SSS-32\cite{huang2018data}     & 2.82 & 74.18                \\ 
%				RBP\cite{zhou2019accelerate}   & 1.78 & 71.50                \\ 
%				RRBP\cite{zhou2019accelerate}  & 1.86 & 73.00                \\ 	
%				GAL-0.1\cite{Lin2019Towards}   & 2.33 & -/71.95              \\ 
%				GAL-0.5\cite{Lin2019Towards}   & 1.58 & -/69.88              \\ 
%				CoGD-0.35 &  -    &  -/-         \\ 
%				CoGD-0.5 &  -    &  -/-         \\ 
%				CoGD-0.75 &  -    &  -/-         \\ 
%				CoGD-0.95 &  -    &  -/-         \\ 
%				\hline\hline
%		\end{tabular}}
%		\label{imagenet_res50}
%		\vspace{-0.3cm}
%	\end{table}	

	For \textbf{MobileNetV2}, the pruning result for MobilnetV2 is summarized in Tab.~\ref{cifar10_res}.  
	Compared to the pre-trained network, CoGD-$0.5$ achieves a \({\rm{0.45}} \times \) FLOPs reduction with a $0.18$\% accuracy drop.
	%With the pruning rate drop, %CoGD-$0.75$ achieves $0.15$\% accuracy increase, and 
	%CoGD-$0.95$ achieves the highest performance $94.83$\% with  \({\rm{0.07}} \times \)  FLOPs reduction. 
	This result indicates that  CoGD is easily employed on efficient networks with depth-wise separable convolution, which is worth exploring in practical applications.

	\subsubsection{Experiments on ILSVRC12 ImageNet}
	For ILSVRC12 ImageNet, we test our CoGD based on  ResNet-50. We train the network with the batch size of $256$ for $60$ epochs. The initial learning rate is set to $0.01$  scaled by $0.1$ over $15$ epochs. Other hyperparameters follow the setting on CIFAR-10. 
	%Unlike the strategy on CIFAR-10, we add the soft mask after both the first and second convolutional layers in each block. 
	The fine-tuning process follows the setting on CIFAR-10 with the initial learning rate $0.00001$.
	
%	\begin{figure}[!t]
%		\centering
%		\includegraphics[scale=0.45]{pic/imagenet.pdf}
%		\caption{\zhuonew{Pruning results of ResNet-50 on ImageNet. Here, B means billion (\({10^9}\)). The pre-trained network of ResNet-50 we use achieves the accuracy 76.24\%\protect\footnotemark.}}
%		\label{losscifar}
%		\vspace{-0.4cm}
%	\end{figure}
%	\footnotetext{The model as our baseline downloads from
%		https://download.pytorch.org/models/resnet50-19c8e357.pth .}
	
	%\subsection{Results on ILSVRC12 ImageNet}
    \zhuonew{Tab.~\ref{imagenet_res50} shows that CoGD achieves the state-of-the-art performance on the ILSVRC12 ImageNet. For ResNet-50, 
    CoGD-$0.5$ further shows a  \({\rm{0.35}} \times \) FLOPs reduction while achieving only a $0.62$\% drop in the accuracy.
    %CoGD-$0.5$ achieves a higher accuracy than SSS-$32$ ($75.69$\% v.s. $74.18$\%) with a smaller FLOPs reduction ($2.67$B v.s. $2.82$B). %With the pruning rate decline, %CoGD-$0.75$ has a  $0.71$\% greater improvement than SSS-$32$ with a similar FLOPs ($2.53B$ v.s. $2.82B$). {Compared to GAL counterparts, CoGD-$0.35$ and CoGD-$0.5$ achieve much better performance than GAL-$0.5$  ($72.07$\% v.s. $69.88$\%) and GAL-$0.1$  ($74.38$\% v.s. $71.95$\%) with similar FLOPs reduction. Furthermore,} 
    %CoGD-$0.5$ further shows a  \({\rm{0.18}} \times \) FLOPs reduction while achieving a $0.54$\% improvement in the accuracy.
    } %Compared other method, Ours have

	\subsubsection{Ablation Study}	
	We use ResNet-18 on CIFAR-10 for ablation study to evaluate the effectiveness of our method. 
	
	\textbf{Effect on CoGD.} We train the pruned network with  \zhuo{and} without CoGD \zhuo{by} using  the same  parameters.
	\zhuo{As shown in Tab.~\ref{ablation_study}}, we obtain an error rate $4.70$\% and a \({\rm{0.51}} \times \) FLOPs reduction 
	with CoGD, compared to the error rate is $5.19$\% and \zhuo{a \({\rm{0.32}} \times \)  FLOPs reduction} without CoGD, which validates the effectiveness of CoGD.
	
	%	\textbf{Effect on $\alpha_W$.}
	%	$\alpha_W$ denotes the  threshold corresponding to the projection proportion $a$. Three projection proportion is tested including  $0.5$, $0.75$ and $0.95$. We project once in the first batch of every epoch and the results with different $\alpha$ are  shown in Tabs.~\ref{cifar10_res} and \ref{imagenet_res50}. %\zhuonew{ We note that CoGD-a means that the projection proportion is $a$\% with an approximate pruning rate of $1-a$\%.}%We note that CoGD-$a$ means that the backtrack rate is $a$\%\zhuo{, and the larger $a$ is, the less FLOPs are reduced.} %\zhuo{controllable}
	\begin{figure}[!t]
		\centering
		\includegraphics[width=0.45\textwidth]{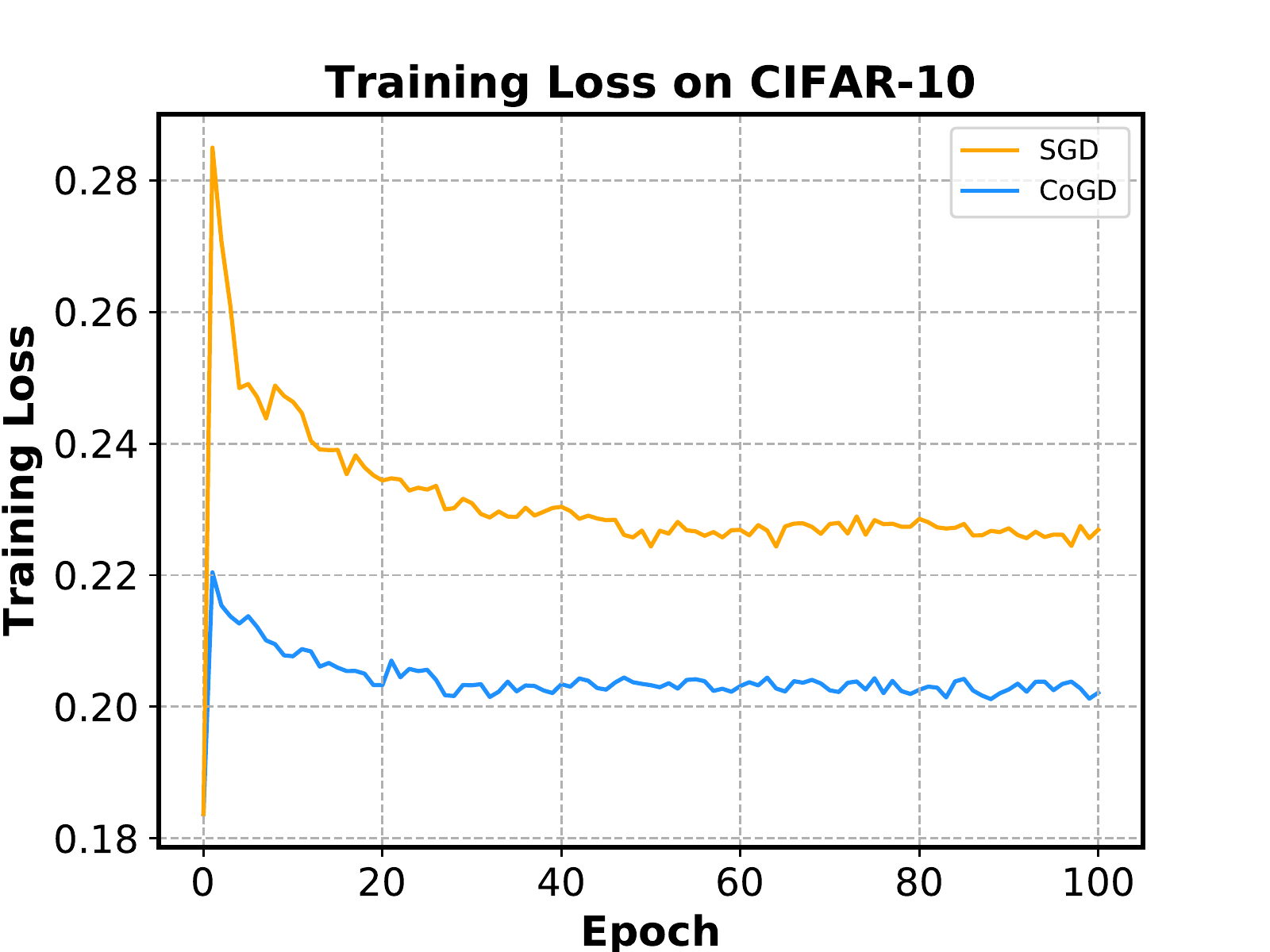}
		\caption{Training loss on CIFAR-10 with CoGD or SGD. }
		\label{convergence}
		\vspace{-0.4cm}
	\end{figure}		

\begin{figure}[t]
		\centering  
		\subfloat{
			\centering
			\label{synchronous}
			\includegraphics[width=0.5\linewidth]{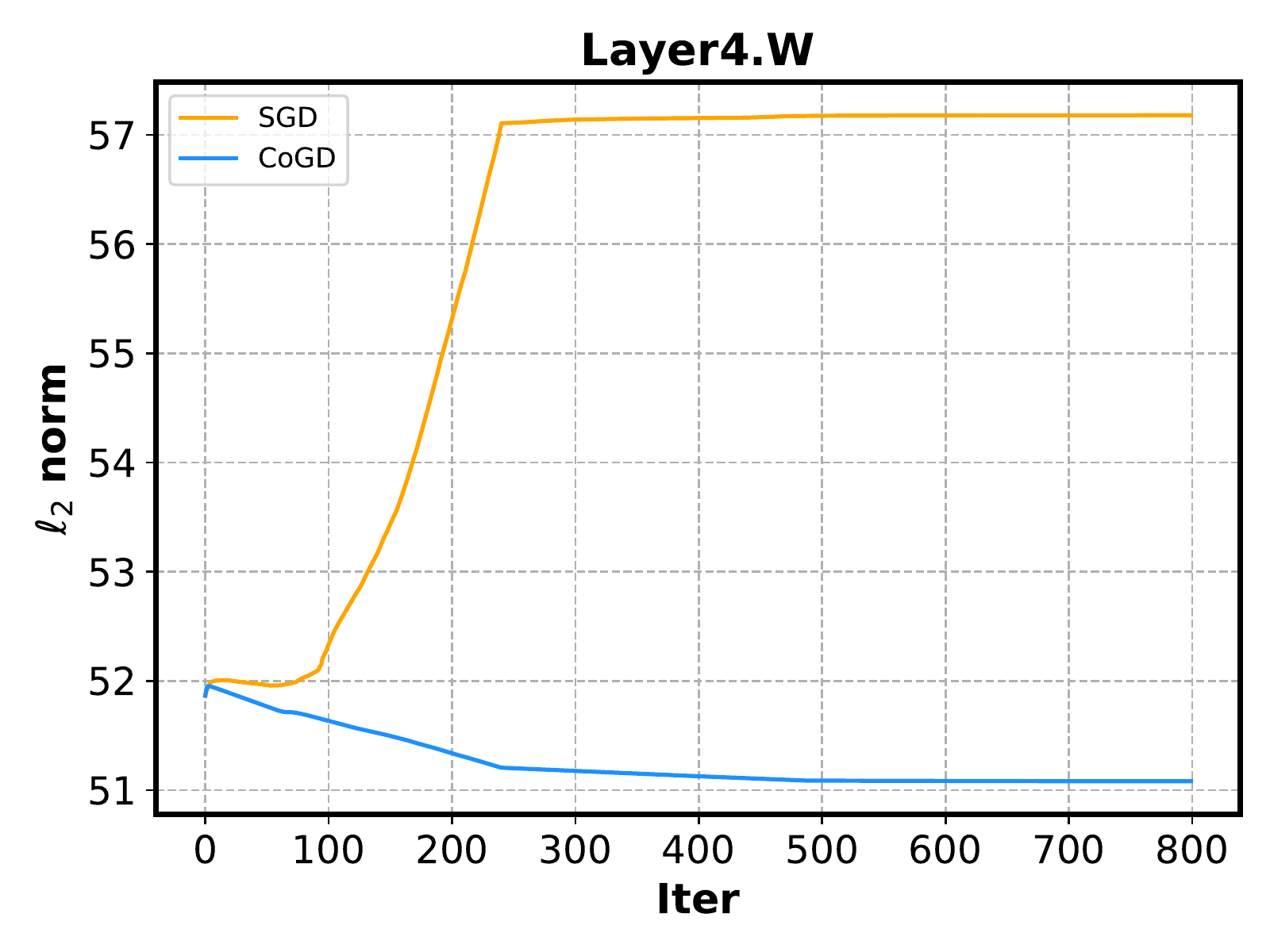}}
		\subfloat{
			\centering
			\label{asynchronous}
			\includegraphics[width=0.5\linewidth]{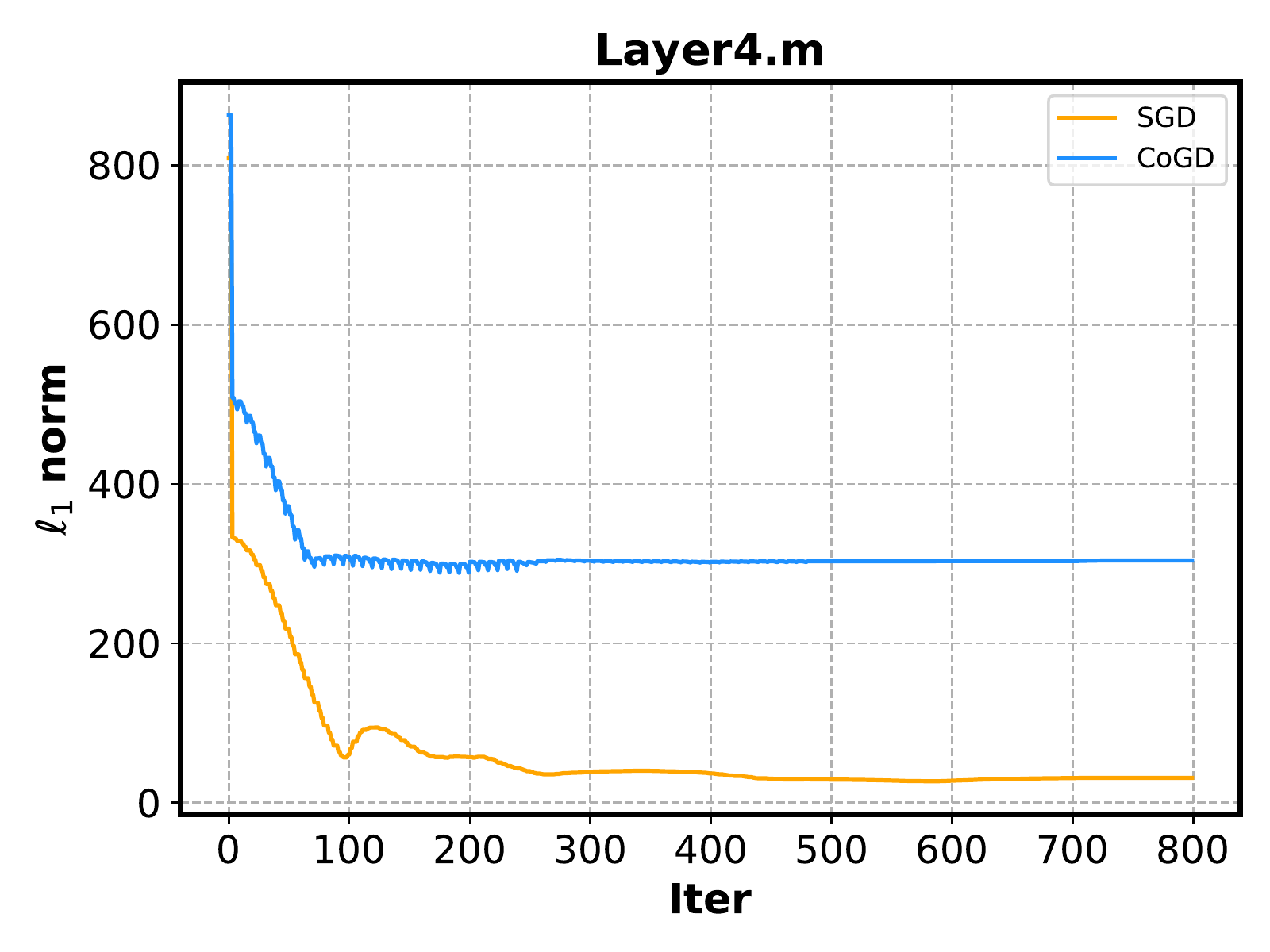}}
		\caption{The left and the right are obtained based on SGD and CoGD-$0.5$ on CIFAR-10.  With CoGD, the two variables converge synchronously while avoiding that either variable quickly gets stuck into local minimum, which validates that CoGD avoids vanishing gradient for the coupled variables.}
		\label{fig:convergence}
		\vspace{-0.4cm}
	\end{figure}

	\textbf{Effect on synchronous convergence.}
	\zhuonew{As is shown in Fig~\ref{convergence}, the training curve shows that the convergence of CoGD is similar to GAL with SGD-based optimization within an epoch, especially for the last epochs when converging in a similar speed. We theoretically derive our method within the gradient descent framework, which provides a solid foundation for the convergence analysis of our method.
	The main differences between SGD and CoGD are twofold: (1) we change the initial point for each epoch; (2) we explore the coupling relationship between the hidden factors to improve a bilinear model within the gradient descent framework. They cause little difference between two methods on the convergence.
	%In Fig.~\ref{fig:convergence}, we show the convergence of two coupled variables from bilinear models with/without CoGD during pruning. We observe that without CoGD, $W$ gets stuck while $\bm$ collapsing suddenly. In contrast, our algorithm with CoGD coordinates the gradient of those two variables and makes them converge synchronously.
	
	In Fig.~\ref{fig:convergence}, we show the convergence in a synchronous manner of the  $4$th
	layer’s variables when pruning CNNs. For better visualization, the learning rate of $\bm$ is enlarged to 100 times. On the curves, we can observe that the two variables converge synchronously while avoiding that either variable quickly gets stuck into local minimum, which validates that CoGD avoids vanishing gradient for the coupled variables.}
	
	\begin{table}[H]
		\centering
		\caption{Pruning results on CIFAR-10 with CoGD or SGD. M means million (\({10^6}\)).}
		\scalebox{0.9}{
			\begin{tabular}{c c c}
				\hline\hline
				Optimizer &Accuracy (\%) &FLOPs / Baseline (M)\\ \hline
				SGD &94.81  &376.12 / 555.43    \\		
				CoGD &95.30 &274.74 / 555.43    \\ \hline\hline
		\end{tabular}}
		\label{ablation_study}
		\vspace{-0.4cm}
	\end{table}

	% 	\begin{figure}[!t]
	
	% 		\includegraphics[width=0.2\textwidth]{pic/cogradient.pdf}
	
	% 		\includegraphics[width=0.2\textwidth]{pic/unknown.pdf}
	% 		\caption{Convergence in a synchronous manner in pruning CNNs based on a bilinear model.}
	% 		\label{cogradient}
	% 		\vspace{-0.4cm}		
	
	% 	\end{figure}
	
	%		\begin{figure}
	%			\centering
	%			\subfigure{
	%				\label{Train Error}
	%			\includegraphics[width=35mm]{pic/cogradient.pdf} }
	%			\hspace{0.5mm} 
	%			\subfigure{
	%				\label{Test Error}
	%			\includegraphics[width=35mm]{pic/unknown.pdf} }
	%			\vspace{-0.1cm}
	%			\caption{Convergence in a synchronous manner in pruning %CNNs based on a bilinear model.}
	%			\vspace{-0.2cm}
	%			\label{fig:loss}
	%		\end{figure}

	\section{Conclusion}
	In this paper, we developed an optimization framework for the bilinear problem with sparsity constraints.
	Since previous gradient based algorithms ignore the coupled relationship between variables and lead to asynchronous gradient descent,
	we introduced CoGD which coordinates the gradient of hidden variables. 
	We applied our algorithm on CSC and CNNs to solve image reconstruction, image inpainting and network pruning. 
	Experiments on different tasks demonstrate that our CoGD outperforms previous algorithms.
	
	\section{Acknowledgements}
	Baochang Zhang is also with Shenzhen Academy of Aerospace Technology, Shenzhen
	100083, China. National Natural Science Foundation of China under Grant 61672079, in part by Supported by Shenzhen Science and Technology Program (No.KQTD2016112515134654). 	
	{\small
		\bibliographystyle{ieee_fullname}
		\bibliography{egbib}
	}
	
\end{document}

%% file: algorithm/algorithm_CSC.tex
	\begin{algorithm}[tb]
	\caption{CoGD for CSC.}
	\label{alg_csc}
	\begin{algorithmic}[1]
		\REQUIRE The  training  dataset; sparisty factor \( \lambda\); hyperparameters such as penalty parameters, threshold $\alpha_{\bA}$, $\alpha_{\bx}$.
		\ENSURE The filters $\ba$ and the sparse feature maps $\bx$.
		\STATE Initialize $\ba^0, \bx^0$ and others
		\REPEAT
		\STATE Use $f_3({\ba_k})$ in Eq.~\ref{eq:proximal_csc} to calculate kernel norm 
		\STATE 
		Use Eq.~\ref{bt_csc} to calculate $\bx$
		\FORALL {$l = 1$ to $L$ epochs}
		%\FORALL {$i $ steps}
		\STATE Kernel Update:\\
		\(\ba^l\) \( \gets \mathop {\arg \min }\limits_{\ba}f_1(\bA\bx)+\sum_{k=1}^K f_3(\ba_k)\) using ADMM with proximal operators  
		\STATE Code Update:\\ 
		\(\bx^l\) \( \gets  \mathop  {\arg \min}\limits_{\bx}f_1(\bA\bx)+\sum_{k=1}^K f_2(\bx_k)\)  using ADMM with proximal operators  
		%Fix $\bd$  and update $\bz$  use Eqs. \ref{optimization}
		%\ENDFOR
		\ENDFOR
		
		\UNTIL{Loss convergence.}
	\end{algorithmic}
\end{algorithm}

%% file: algorithm/algorithm_pruning.tex
	\begin{algorithm}[tb]
	\caption{CoGD for  Pruning CNNs in Bilinear Modeling.}
	\label{alg_gal}
	\begin{algorithmic}[1]
		\REQUIRE
		The training dataset; the pre-trained network with weights $W_B$;
		sparisty factor \( \lambda\);
		hyperparameters such as learning rate, weight decay, threshold $\alpha_W$, $\alpha_{\bm}$.
		\ENSURE
		The pruning network.
		%The convolutional result $F^{l+1}$ using circulant binary convolutional networks.
		\STATE Initialize \( W_G = W_B\), \(\bm \sim N(0,1)\);
		\REPEAT
		\STATE Use $R(W_G)$ in Eq.~\ref{problem_pruning} to obtain the norms
		\STATE Use Eq.~\ref{bt_pruning} to calculate the new soft mask $\hat{\bm}$			
		\FORALL {$l = 1$ to $L$ epochs}
		
		\FORALL {$i $ steps}
		\STATE Fix $W_G$  and update $W_D$ using Eq. \ref{optimization}
		\ENDFOR
		\FORALL {$j $ steps}
		\STATE Fix $W_D$  and update $W_G$  using Eq. \ref{optimization}
		\ENDFOR
		\ENDFOR
		
		\UNTIL{Loss convergence.}
	\end{algorithmic}
\end{algorithm}